\documentclass[11pt,a4paper]{article}
\usepackage[hyperref]{acl2019}
\usepackage{times}
\usepackage{latexsym}
\usepackage{comment}
\usepackage{todonotes}
\usepackage{graphicx}
\usepackage{graphics}
\usepackage{amsmath}
\usepackage{url}
\usepackage{enumitem}
\usepackage{color}
\usepackage{url}
\usepackage{booktabs}
\usepackage{multirow}
\usepackage{todonotes}
\aclfinalcopy 

\title{Towards Best Experiment Design for Evaluating Dialogue System Output}
\author{Sashank Santhanam  and Samira Shaikh\\
  Computer Science \\
  University of North Carolina at Charlotte \\
  Charlotte, NC, USA\\
  \texttt{\{ssantha1,sshaikh2\}@uncc.edu} \\}

\date{}

\begin{document}
\maketitle
\begin{abstract}
To overcome the limitations of automated metrics (e.g. BLEU, METEOR) for evaluating dialogue systems, researchers typically use human judgments to provide convergent evidence. While it has been demonstrated that human judgments can suffer from the inconsistency of ratings, extant research has also found that the \textit{design} of the evaluation task affects the consistency and quality of human judgments. We conduct a between-subjects study to understand the impact of four experiment conditions on human ratings of dialogue system output. In addition to discrete and continuous scale ratings, we also experiment with a novel application of Best-Worst scaling to dialogue evaluation. Through our systematic study with 40 crowdsourced workers in each task, we find that using continuous scales achieves more consistent ratings than Likert scale or ranking-based experiment design. Additionally, we find that factors such as time taken to complete the task and no prior experience of participating in similar studies of rating dialogue system output \textit{positively} impact consistency and agreement amongst raters. 
\end{abstract}

\section{Introduction and Related Work} 
A tremendous amount of recent research has focused on approaches towards generating responses for conversations in an open-domain setting \cite{radford2019language,xing2018hierarchical,wolf2019transfertransfo}. An equally challenging task for natural language generation systems is evaluating the quality of the generated responses. 
Evaluation of generated output is typically conducted using a combination of crowdsourced human judgments and automated metrics adopted from machine translation and text summarization \cite{dialogue-eval,novikova-etal-2017-need}. However, studies conducted by Liu \emph{et al.}\citeyearpar{dialogue-eval} and Novikova \emph{et al.} \citeyearpar{novikova-etal-2017-need} show that the automated metrics have poor correlation with human judgments. Despite their shortcomings, automated metrics like BLEU, ROUGE, and METEOR are used due to a lack of alternative metrics. This puts a major imperative on obtaining high-quality crowdsourced human judgments. 
Previous research which employs crowdsourced judgments has focused on metrics including \textit{ease of answering}, \textit{information flow} and \textit{coherence} \cite{Li-reinforce, dziri2018augmenting}, \textit{naturalness} \cite{asghar2018affective}, \textit{interestingness} \cite{asghar-etal-2017-deep,santhanam2019survey}, \textit{fluency} or \textit{readability} \cite{zhang-etal-2018-personalizing}, \textit{engagement} \cite{venkatesh2018evaluating}. While experiment designs primarily use Likert scales, Belz and Kow \citeyearpar{belz2010comparing} argue that discrete scales, such as the Likert scales, can be unintuitive and certain individuals may avoid extreme values in their judgments. Prior research has also shown that use of continuous scales is more viable for language evaluation \cite{novikova-etal-2018-rankme,belz2011discrete}. Such evidence places more emphasis on a careful study towards obtaining reliable and consistent human ratings for dialogue evaluation. 

To address this research problem, we focus on a systematic comparison of four experimental conditions by incorporating \textbf{\textit{continuous, relative}} and \textbf{\textit{ranking scales}} for obtaining crowdsourced human judgments. In this initial study, we evaluate the use of two metrics:  \textbf{\textit{Readability}} and \textbf{\textit{Coherence}}. 

Our key findings are: 
\begin{enumerate} [nolistsep,noitemsep]
    \item Use of Likert scales results in the lowest inter-rater consistency and agreement when compared to other experiment conditions
    \item Use of continuous scales results in higher inter-rater consistency and agreement
    \item Raters who have no prior experience in evaluating dialogue system output have greater inter-rater consistency and agreement than do those who have previously participated in such rating tasks. 
\end{enumerate}
Our findings have the potential to help the research community in the design of their evaluation tasks to obtain higher quality human judgments for natural language generation output.

\section{Data and Models}
We used the Reddit conversation corpus to train our models. The Reddit conversation corpus, made available by Dziri \emph{et al.} \citeyearpar{dziri2018augmenting}, consists of data extracted from 95 top-ranked subreddits that discuss various topics such as sports, news, education and politics. The corpus contains 9M training examples, 500K development dialogues and 400K dialogues as test data.\footnote{\url{https://github.com/nouhadziri/THRED}} We trained three models on the Reddit conversational dataset described below. All the pre-trained models and supporting analysis code along with user study data are available at \url{https://www.github.com/sashank06/INLG_eval}. The models trained for this study include:

    $\bullet$ \textbf{Seq2Seq:} Simple encoder-decoder model with attention mechanism \cite{bahdanau2014neural}
    
    $\bullet$ \textbf{HRED:} \textit{\textbf{Hierarchical Encoder-Decoder}} \cite{serban2016building} which incorporates an utterance and intra-utterance layer to model context.
    
    $\bullet$ \textbf{THRED:} \textit{\textbf{Topic Augmented Hierarchical Encoder-Decoder}} \cite{dziri2018augmenting} which uses topic words along with a hierarchical encoder-decoder to produce a response.

\section{Metrics}
\label{metrics}
For this initial study, we focus on two metrics, readability and coherence. These metrics are among those essential to evaluate the quality of generated responses \cite{novikova-etal-2017-need,dziri-etal-2019-evaluating-coherence}. We describe an automated method to compute each metric.

\textbf{Readability} or Fluency measures the linguistic quality of text and helps quantify the difficulty of understanding the text for a reader  \cite{gatt2018survey,novikova-etal-2017-need}. We use the Flesch Reading Ease (FRE) \cite{kincaid1975derivation} that counts the number of words, syllables and sentences in the text.\footnote{\url{https://bit.ly/1IZ0FG4}} Higher readability scores indicate that utterance is easier to read and comprehend.
    
\textbf{Coherence} measures the ability of the dialogue system to produce responses consistent with the topic of conversation \cite{venkatesh2018evaluating}. To calculate coherence, we use the method proposed by Dziri \emph{et al.} \citeyearpar{dziri2018augmenting}. This metric computes the cosine similarity on embedding vectors of generated response and target while accounting for dull and generic responses through a penalty factor.

To overcome the issue of dull and generic responses, Dziri \emph{et al.}  \citeyearpar{dziri2018augmenting} induce a penalty factor which takes into account 
    \begin{equation}
        P = 1 + \log \frac{2+L'}{2+L''}
    \end{equation}
    where $L'$ indicates the length of response after dropping stop words and punctuation and $L''$ indicates the length of non-dull parts of the response after dropping stop words. The penalized semantic similarity (SS) score is then calculated as:
    \begin{equation}
        SS(utt_{i,j},resp_i) = P \times ( 1-cos(utt_{i,j},resp_i)
    \end{equation}
    where $i$ represents the index of the dialogue in the dataset and $j$ denotes index of the utterance in the conversation history.

\section{Experiment Designs}
In our study, we use three well-known question types of Likert Scale, Magnitude Estimation and Best-Worst Ranking. We chose these questions types to investigate as these are commonly used across various language evaluation tasks \cite{belz2011discrete,asghar2018affective,novikova-etal-2018-rankme,kiritchenko-mohammad-2017-best} . With the help of these three types of questions, we design four rating procedures that are explained below.



\textbf{Likert Scale (LS)}: is typically used in experiments for crowdsourcing human evaluation of dialogue systems \cite{asghar2018affective,lowe-etal-2017-towards}. In our experiment, we ask the raters to rate the generated responses on a 6-point scale, following Novikova \emph{et al.} \citeyearpar{novikova-etal-2018-rankme} (where 1 is the lowest and 6 is the highest on the metrics of readability and coherence). 

\textbf{Rank-Based Magnitude Estimation (RME)}: Prior research by Belz and Kow \citeyearpar{belz2011discrete} demonstrates through six separate experiments that continuous scales are more viable and offer distinct advantages over discrete scales in evaluation tasks. Recently, Novikova \emph{et al.} \citeyearpar{novikova-etal-2018-rankme} adopted magnitude estimation by providing the rater with a \textit{standard value} for a reference sentence to evaluate output from goal-oriented systems. Following Novikova \emph{et al.} \citeyearpar{novikova-etal-2018-rankme}, we also set the value of the standard (reference utterance) as 100 since the reference utterance was produced by humans and is considered as gold-standard. The crowd-sourced workers are asked to provide a  score relative to 100 (from 0 to 999) for three system-generated outputs. 


\textbf{Biased Magnitude Estimation (BME)}: Our third experiment design is biased magnitude estimation (BME). The main difference between RME and BME method is that the standard value we provide for the reference utterance is not uniformly set to 100 for all examples, but instead calculated by automated methods (explained in Section \ref{metrics}). Our motivation to do so is to understand if \textbf{anchoring bias} may affect the ratings when judgments are made relative to a fixed value (100) or relative to a value calculated by automated means. Anchoring bias is the tendency to rely too heavily on one piece of information offered (the ``anchor'', in this case, the number 100) when making decisions \cite{kahneman201636}. 

\textbf{Best-Worst Scaling (BWS)}: Our last experiment condition is best-worst scaling (BWS) in which raters are asked to rank the generated responses in order of best to worst on both metrics (readability and coherence). This approach has previously been used to estimate emotion intensity and has been demonstrated to produce high quality and consistent judgments from humans \cite{kiritchenko-mohammad-2017-best}.

Each task includes 50 randomly sampled conversations from the test set in our corpus along with generated responses from the three models and the ground truth (reference utterance). For each task, we collected ratings from 40 workers with Master qualifications through Amazon Mechanical Turk. 

    
    
    


\section{Experiment Results}
We organize our findings along five main research questions (RQs) outlined in this section. In the following section, we report on statistical significance using two-way ANOVAs on the between-subject ratings across the four experiment conditions (Tables~\ref{reliability-all}\textendash ~\ref{q2-no}).

\textbf{RQ1: What is the effect of experiment design on the reliability on human ratings?}
We use intra-class correlation (ICC) to measure the reliability across multiple raters \cite{shrout1979intraclass,landis1977measurement}. To compare the scores obtained from magnitude estimation experiments to the ratings from the task using discrete Likert scales, we perform a normalization of the magnitude estimation scores on a logarithmic scale as suggested by Bard \emph{et al.} \citeyearpar{bard1996magnitude}.
\begin{table}[t]
\centering
\small
\begin{tabular}{@{}p{0.7cm}p{1.2cm}p{0.8cm}p{0.8cm}p{0.8cm}p{0.8cm}@{}}
\toprule
 &  & \multicolumn{1}{c}{\textbf{Likert}} & \multicolumn{1}{c}{\begin{tabular}[l]{@{}l@{}} \textbf{RME} \end{tabular}} & \multicolumn{1}{c}{\textbf{\begin{tabular}[l]{@{}l@{}} \textbf{BME} \end{tabular}}} & \multicolumn{1}{c}{\begin{tabular}[l]{@{}l@{}} \textbf{BWS} \end{tabular}} \\ \midrule

\multirow{2}{*}{ICC-C} & Readability & 0.75 & 0.95$\dagger$ & 0.83 & 0.75\\ \cmidrule(l){2-6} 
 & Coherence & 0.83 & 0.92 & 0.81 & 0.80 \\ \midrule
 
\multirow{2}{*}{ICC-A} & Readability & 0.59& 0.95$\dagger$ & 0.83 & 0.75\\ \cmidrule(l){2-6} 
 & Coherence & 0.77 & 0.92 & 0.81  & 0.80\\ \bottomrule

\end{tabular}
\caption{ICC scores on the metrics of readability and coherence for each experiment design. All values are statistically significant  p-value\textless{0.001} except those indicated by $\dagger$. n$=$40 for all four designs.}
\label{reliability-all}
\end{table}

Table \ref{reliability-all} represents the ICC scores on consistency (ICC-C) and agreement (ICC-A) for our four experiment tasks. We observe that use of Magnitude Estimation with anchors (RME or BME) results in more reliable ratings than using Likert Scale or using Best-Worst ranking (BWS). This result is consistent with prior research by Novikova \emph{et al.} \citeyearpar{novikova-etal-2018-rankme} and Belz and Kow \citeyearpar{belz2011discrete}. 

\textbf{RQ2: Does time taken to complete the survey influence reliability of the rankings?} 
To analyze RQ2, we calculated the total time spent by each participant from the start to the end of the experiment. We found that BME task had longest on average time to completion (43 minutes), followed by RME (42.8 minutes) and Likert scale (33 minutes; Best-Worst ranking had shortest average completion time (32.5 minutes). We then test the hypothesis that raters who spent longer than average time on the task would be more reliable in their ratings than those who completed in less than average time. Table \ref{above-mean} represents the ICC scores for raters who spent higher than average time for the task, while Table \ref{below-mean} represents scores for raters who spent less than average time. Surprisingly, we find that consistency and agreement among raters who spend less than average time is higher than those who spend more time, for the Likert, BME or BWS experiment designs. When using the RME design, raters who spend more time have higher consistency and agreement.

\begin{table}[!htbp]
\centering
\small
\begin{tabular}{@{}p{0.7cm}p{1.2cm}p{0.8cm}p{0.8cm}p{0.8cm}p{0.8cm}@{}}
\toprule
 &  & \multicolumn{1}{c}{\begin{tabular}[l]{@{}l@{}} \textbf{Likert} \\ (n=15) \end{tabular}} & \multicolumn{1}{c}{\begin{tabular}[l]{@{}l@{}} \textbf{RME} \\ (n=16) \end{tabular}} & \multicolumn{1}{c}{\begin{tabular}[l]{@{}l@{}} \textbf{BME} \\ (n=15) \end{tabular}} & \multicolumn{1}{c}{\begin{tabular}[l]{@{}l@{}} \textbf{BWS} \\(n=16) \end{tabular}} \\ \midrule

\multirow{2}{*}{ICC-C} & Readability & 0.58& 0.93 & 0.51 & 0.62\\ \cmidrule(l){2-6} 
 & Coherence & 0.74 & 0.85 & 0.55 & 0.64 \\ \midrule
 
\multirow{2}{*}{ICC-A} & Readability & 0.52 & 0.93 & 0.51 & 0.62\\ \cmidrule(l){2-6} 
 & Coherence & 0.69 & 0.86& 0.56  & 0.64\\ \bottomrule

\end{tabular}
\caption{ICC scores when participants spend \textbf{above average time}. All values in this table are statistically significant with p-value\textless{0.001}}
\label{above-mean}
\end{table}

\begin{table}[h]
\centering
\small
\begin{tabular}{@{}p{0.7cm}p{1.2cm}p{0.8cm}p{0.8cm}p{0.8cm}p{0.8cm}@{}}
\toprule
 &  & \multicolumn{1}{c}{\begin{tabular}[l]{@{}l@{}} \textbf{Likert} \\ (n=25) \end{tabular}} & \multicolumn{1}{c}{\begin{tabular}[l]{@{}l@{}} \textbf{RME} \\ (n=24) \end{tabular}} & \multicolumn{1}{c}{\begin{tabular}[l]{@{}l@{}} \textbf{BME} \\ (n=25) \end{tabular}} & \multicolumn{1}{c}{\begin{tabular}[l]{@{}l@{}} \textbf{BWS}\\(n=24) \end{tabular}} \\ \midrule

\multirow{2}{*}{ICC-C} & Readability & 0.61 & 0.88 & 0.81 & 0.65\\ \cmidrule(l){2-6} 
 & Coherence & 0.66 & 0.85 & 0.75 & 0.76 \\ \midrule
 
\multirow{2}{*}{ICC-A} & Readability & 0.36 & 0.88 & 0.81 & 0.66\\ \cmidrule(l){2-6} 
 & Coherence & 0.55 & 0.85 & 0.75  & 0.76\\ \bottomrule

\end{tabular}
\caption{ICC scores when participants spend \textbf{below average time}. All values in this table are statistically significant with p-value\textless{0.001}}
\label{below-mean}
\end{table}

\textbf{RQ3: Does prior experience of evaluating dialogue system output or engaging with conversational agents affect reliability of rankings?}
We asked each rater two additional questions at the end of the task. The questions asked raters to indicate whether or not they had prior experience taking part in studies (a) to evaluate dialogue system output; and (b) to engage with a conversational agent.

Tables \ref{q1-yes} and \ref{q1-no} show how reliable the ratings from the participants based on their prior experience of taking part in studies about evaluating conversational response. We find that participants who have not taken part in prior studies are more consistent and have a higher agreement score than participant who have prior experience. These results are also validated by Tables \ref{q2-yes} and \ref{q2-no} which shows that participants with no prior experience of engaging with conversational agents are more consistent and reliable.

\begin{table}[h]
\centering
\small
\begin{tabular}{@{}p{0.7cm}p{1.2cm}p{0.8cm}p{0.8cm}p{0.8cm}p{0.8cm}@{}}
\toprule
 &  & \multicolumn{1}{c}{\begin{tabular}[l]{@{}l@{}} \textbf{Likert} \\ (n=15) \end{tabular}} & \multicolumn{1}{c}{\begin{tabular}[l]{@{}l@{}} \textbf{RME} \\ (n=7) \end{tabular}} & \multicolumn{1}{c}{\begin{tabular}[l]{@{}l@{}} \textbf{BME} \\ (n=18) \end{tabular}} & \multicolumn{1}{c}{\begin{tabular}[l]{@{}l@{}} \textbf{BWS}\\(n=13) \end{tabular}} \\ \midrule

\multirow{2}{*}{ICC-C} & Readability & 0.45 & 0.37 & 0.51 & 0.54\\ \cmidrule(l){2-6} 
 & Coherence & 0.38 & 0.48 & 0.55 & 0.63 \\ \midrule
 
\multirow{2}{*}{ICC-A} & Readability & 0.35 & 0.38 & 0.52 & 0.55\\ \cmidrule(l){2-6} 
 & Coherence & 0.32 & 0.49 & 0.55  & 0.63\\ \bottomrule

\end{tabular}
\caption{ICC scores when participants \textbf{have} prior experience evaluating dialogue system output. All values statistically significant at p-value\textless{0.001}.}
\label{q1-yes}
\end{table}

\begin{table}[h]
\centering
\small
\begin{tabular}{@{}p{0.7cm}p{1.2cm}p{0.8cm}p{0.8cm}p{0.8cm}p{0.8cm}@{}}
\toprule
 &  & \multicolumn{1}{c}{\begin{tabular}[l]{@{}l@{}} \textbf{Likert} \\ (n=25) \end{tabular}} & \multicolumn{1}{c}{\begin{tabular}[l]{@{}l@{}} \textbf{RME} \\ (n=33) \end{tabular}} & \multicolumn{1}{c}{\begin{tabular}[l]{@{}l@{}} \textbf{BME} \\ (n=22) \end{tabular}} & \multicolumn{1}{c}{\begin{tabular}[l]{@{}l@{}} \textbf{BWS}\\(n=27) \end{tabular}} \\ \midrule

\multirow{2}{*}{ICC-C} & Readability & 0.71 & 0.95$\dagger$ & 0.83 & 0.70\\ \cmidrule(l){2-6} 
 & Coherence & 0.82 & 0.92 & 0.76 & 0.72 \\ \midrule
 
\multirow{2}{*}{ICC-A} & Readability & 0.50 & 0.95$\dagger$ & 0.83 & 0.70\\ \cmidrule(l){2-6} 
 & Coherence & 0.75 & 0.92 & 0.77  & 0.72\\ \bottomrule

\end{tabular}
\caption{ICC scores when participants \textbf{do not have} prior experience evaluating dialogue system output. All values statistically significant at p-value\textless{0.001} except those indicated by $\dagger$.}
\label{q1-no}
\end{table}

\begin{table}[h]
\centering
\small
\begin{tabular}{@{}p{0.7cm}p{1.2cm}p{0.8cm}p{0.8cm}p{0.8cm}p{0.8cm}@{}}
\toprule
 &  & \multicolumn{1}{c}{\begin{tabular}[l]{@{}l@{}} \textbf{Likert} \\ (n=18) \end{tabular}} & \multicolumn{1}{c}{\begin{tabular}[l]{@{}l@{}} \textbf{RME} \\ (n=11) \end{tabular}} & \multicolumn{1}{c}{\begin{tabular}[l]{@{}l@{}} \textbf{BME} \\ (n=23) \end{tabular}} & \multicolumn{1}{c}{\begin{tabular}[l]{@{}l@{}} \textbf{BWS}\\(n=18) \end{tabular}} \\ \midrule

\multirow{2}{*}{ICC-C} & Readability & 0.46 & 0.69 & 0.60 & 0.57\\ \cmidrule(l){2-6} 
 & Coherence & 0.44 & 0.65 & 0.62 & 0.67 \\ \midrule
 
\multirow{2}{*}{ICC-A} & Readability & 0.37 & 0.69 & 0.61 & 0.57\\ \cmidrule(l){2-6} 
 & Coherence & 0.38 & 0.65 & 0.62  & 0.67\\ \bottomrule

\end{tabular}
\caption{ICC scores when participants \textbf{have} prior experience engaging with conversational agents. All values statistically significant at p-value\textless{0.001}.}
\label{q2-yes}
\end{table}

\begin{table}[!htbp]
\centering
\small
\begin{tabular}{@{}p{0.7cm}p{1.2cm}p{0.8cm}p{0.8cm}p{0.8cm}p{0.8cm}@{}}
\toprule
 &  & \multicolumn{1}{c}{\begin{tabular}[l]{@{}l@{}} \textbf{Likert} \\ (n=22) \end{tabular}} & \multicolumn{1}{c}{\begin{tabular}[l]{@{}l@{}} \textbf{RME} \\ (n=29) \end{tabular}} & \multicolumn{1}{c}{\begin{tabular}[l]{@{}l@{}} \textbf{BME} \\ (n=17) \end{tabular}} & \multicolumn{1}{c}{\begin{tabular}[l]{@{}l@{}} \textbf{BWS}\\(n=22) \end{tabular}} \\ \midrule

\multirow{2}{*}{ICC-C} & Readability & 0.70 & 0.95$\dagger$ & 0.84 & 0.67\\ \cmidrule(l){2-6} 
 & Coherence & 0.82 & 0.91 & 0.76 & 0.68 \\ \midrule
 
\multirow{2}{*}{ICC-A} & Readability & 0.48 & 0.95$\dagger$ & 0.84 & 0.67\\ \cmidrule(l){2-6} 
 & Coherence & 0.75 & 0.91 & 0.76  & 0.68\\ \bottomrule

\end{tabular}
\caption{ICC scores when participants \textbf{do not have} prior experience engaging with conversational agents. All values statistically significant at p-value\textless{0.001} except those indicated by $\dagger$.}
\label{q2-no}
\end{table}

\textbf{RQ4: How well do automated methods to calculate readability and coherence correlate with human ratings?} We report on correlation between readability and coherence scores that are calculated using automated methods (outlined in Section~\ref{metrics}) with the human ratings in Table~\ref{automated}. Readability scores were computed using the Flesh Reading Ease \cite{kincaid1975derivation} and coherence scores were computed based on method proposed by Dziri \emph{et al.} \citeyearpar{dziri2018augmenting}. We observe that the automated metrics for Readability \cite{kincaid1975derivation} and Semantic Similarity \cite{dziri2018augmenting} show low correlation to human judgments ratings.

\begin{table}[!htbp]
\small
\centering
\begin{tabular}{@{}lllll@{}}
\toprule
\textbf{} & \textbf{Likert} & \textbf{RME} & \textbf{BME} & \textbf{BWS} \\ \midrule
 & \multicolumn{4}{c}{Automated Metric} \\ \midrule
Readability & 0.26 & -0.11 & -0.12 & -0.06 \\ \midrule
Coherence & -0.12 & -0.13 & -0.11 & 0.01 \\ \bottomrule
\end{tabular}
\caption{Spearman correlation between the ratings obtained from the automated metrics to human ratings.}
\label{automated}
\end{table}

\textbf{RQ5: Is there any correlation between ratings of readability and coherence for each of the four experiment conditions?} To evaluate whether there is any correlation between the ratings obtained for readability and coherence through of four experimental designs, we report the Spearman correlation values in Table \ref{correaltion_within}. We find that there is high correlation between the human ratings of readability and coherence obtained through RME and BME (statistically significant). One likely factor affecting  correlation may be anchoring bias towards the fixed value of the standard utterance provided in RME (100) and reference value provided in BME. We aim to investigate this further in future work.

\begin{table}[h]
\small
\centering
\begin{tabular}{@{}lllll@{}}
\toprule
 & \textbf{Likert} & \textbf{RME} & \textbf{BME} & \textbf{BWS} \\ \midrule
 & \multicolumn{4}{c}{Readability} \\ \midrule
Coherence & 0.1 & 0.79*** & 0.77*** & 0.5*** \\ \bottomrule
\end{tabular}
\caption{Spearman correlation between the ratings of readability and coherence obtained on four different experiment designs. *** p-value\textless{0.001}}
\label{correaltion_within}
\end{table}

\section{Conclusion and Future Work}
In this paper, we present our work on designing a systematic experiment with four experiment conditions to evaluate the output of dialogue systems. Different from prior work where a similar study was conducted with output from goal-oriented systems \cite{novikova-etal-2018-rankme}, our study focuses on evaluating output in open-domain situations. Consistent with prior findings, metrics calculated using automated methods \cite{dziri-etal-2019-evaluating-coherence} were found to have a negative correlation with human judgments (c.f. Table~\ref{automated}). This finding points to the need for more effective automated metrics.

We find that that use of continuous scales to obtain crowdsourced ratings provides more consistent and reliable ratings than ratings obtained through Likert scales or Best-Worst scaling. This finding is consistent with prior work conducted by Novikova \emph{et al.} \citeyearpar{novikova-etal-2018-rankme}. Novel in our study was the testing of the Best-Worst scaling method to evaluate responses against one another. Although the Best-Worst scaling method has been shown to be effective in obtaining crowdsourced ratings of emotions \cite{kiritchenko-mohammad-2017-best}, we did not find it to be effective in this study. We aim to investigate further whether this finding can be reproduced in a different experiment.

Further, we were able to identify the effects of time taken to complete the task on rating reliability. We find that workers who spent less than average time on the task had higher consistency (for the Likert, BME and BWS experiment conditions) than did the workers who spent more than average time. This finding is counter-intuitive, we expect that spending more time would positively impact inter-rater consistency. Our first step in the analysis of the effects of time taken on reliability included analyzing data from workers who spent more or less than average time, which offers admittedly a limited perspective; an interesting next step would be to more thoroughly study the effects of time taken on reliability by taking into account the full distribution of the time spent data. 

We also find that \textit{lack of} prior experience of evaluating open-domain dialogue system output results in more reliable ratings. One potential explanation for this could be that workers may have pre-conceived notions based on their past experience. One limitation of our current study is that although we had output from three separate models, we conducted the study using data from one corpus. Reproducing our findings across additional corpora, additional metrics and other experiment designs would help substantiate these findings further. An analysis of the interaction effects between independent variables such as time taken and prior experience would also help strengthen the findings of our study.

By using a larger sample size (n$=$40), we are able to make claims about statistical significance across experiment conditions. In future work, we plan to evaluate the impact of cognitive biases such as anchoring and confirmation bias in-depth and how it affects consistency and reliability along with testing continuous scale ratings with no reference value.

 \section*{Acknowledgments}
This work was supported by the Defense Advanced Research Projects Agency (DARPA) under Contract No FA8650-18-C-7881. All statements of fact, opinion or conclusions contained herein are those of the authors and should not be construed as representing the official views or policies of AFRL, DARPA, or the U.S. Government. We thank the anonymous reviewers for the helpful feedback.

\bibliography{acl2019}
\bibliographystyle{acl_natbib}

\end{document}